\pdfoutput=1

\documentclass[11pt]{article}

\usepackage{ACL2023}

\usepackage{times}
\usepackage{latexsym}
\usepackage{pdflscape}

\usepackage[T1]{fontenc}

\usepackage[utf8]{inputenc}

\usepackage{microtype}

\usepackage{inconsolata}

\usepackage{tabularx}
\usepackage{longtable}
\usepackage{graphicx}
\usepackage{verbatim}

\newcommand{\ti}{{ type I }}
\newcommand{\tii}{{ type II }}
\newcommand{\tiii}{{ type III }}

\title{Grammatical information \\ in BERT sentence embeddings as two-dimensional arrays}

\author{ Vivi Nastase \and Paola Merlo
\vspace{1.5mm} \\
Department of Linguistics \\ University of Geneva \\ \texttt{Paola.Merlo@unige.ch, vivi.a.nastase@gmail.com} 
}

\begin{document}
\maketitle
\begin{abstract}

Sentence embeddings induced with various transformer architectures encode much semantic and syntactic information in a distributed manner in a one-dimensional array. We investigate whether specific grammatical information can be accessed in these distributed representations. Using data from a task developed to test rule-like generalizations, our experiments on detecting subject-verb agreement yield several promising results. First, we show that while the usual sentence representations encoded as one-dimensional arrays do not easily support extraction of rule-like regularities, a two-dimensional  reshaping of these vectors allows various learning architectures to access such information. Next, we show that various architectures can detect patterns in these two-dimensional reshaped sentence embeddings and successfully learn a model based on smaller amounts of simpler training data, which performs well on more complex test data. This indicates that current sentence embeddings contain information that is regularly distributed, and which can be captured when the embeddings are reshaped into higher dimensional arrays. Our results cast light on representations produced by language models and help move towards developing few-shot learning approaches.
\end{abstract}

\section{Introduction}

Transformer-based models have taken the NLP world, and not only, by storm in recent years. They have even reached super-human performance on standard benchmarks such as SuperGLUE \cite{wang-ea2019} and SQuAD \cite{rajpurkar-etal-2018-know}, and the output of GPT-* and chatGPT are often worryingly difficult to distinguish from human-produced data \cite{marcus2022,susnjak2022}. Considering such performance, the expectations are high that the word and sentence representations produced by transformer-based architectures can also be useful for finer-grained tasks, such as those that target specific grammatical phenomena. 

Long-distance agreement, a specific and simple grammatical phenomenon, is often used to test the syntactic abilities of deep neural networks \cite{linzen2016,gulordava2018,goldberg2019,linzen2021}. Long-distance agreement tests are usually framed as a prediction task: whether the model gives higher probability to the correct form of the verb, despite intervening {\it attractors} -- nouns appearing between the subject and the verb. This decision is somehow implicit: the language model gives a prediction based on the data it has been built on. 

In this paper, we investigate what happens when we target subject-verb agreement explicitly in BERT sentence embeddings for detecting subject-verb agreement in French sentences. We work with raw BERT sentence embeddings -- not fine-tuned for specific tasks -- and investigate how this specific information is encoded in the distributed representations. More specifically, we ask: Can we detect subject-verb agreement in sentence embeddings? And can we manipulate the raw sentence embeddings to make this phenomenon more easy to detect?
To address these questions, we adopt the framework and task definition for rule-like learning described in \citet{merlo2021}. 
In this framework, learning a given linguistic property or rule is formulated as a task of implicit rule detection in a multiple-choice setup, based on input sequences that share the target property. For the problem under investigation here, each sequence consists of sentences that share subject-verb agreement, but have different distances between the subject and the verb, different clause structures, and different agreement patterns. 

We show that BERT sentence embeddings encoded as a one-dimensional array are only successful at detecting subject-verb agreement when provided with large amounts of data. Reshaping the embeddings to two-dimensional arrays, and combining these with VAE-based architectures, allows a system to detect better the shared patterns in the input sequences, while relying on much smaller amounts of simpler data.
These results open up new avenues of exploration for few-shot learning \citep{feifei-2006,brown-ea2020-neurips,schick-schutze-2021-exploiting}. They also support further analyses of more disentangled representations, those representations that encode underlying rule-like generalisations, typical of human knowledge representation, but not of neural networks \citep{sable-meyer-ea2021}.
The contributions of this paper are:

\begin{enumerate}
    \item We show that that there are higher-dimension patterns that encode syntactic phenomena in BERT sentence embeddings,  beyond the one-dimensional array representation that is readily available from the transformer output.
    \item We show that, used together with VAE-based architectures, two-dimensional reshapings of these representations facilitate the discovery of patterns that encode specific targeted grammatical phenomena.
    \item We show that, through the 2D-ed representations, we get better access to encoded patterns, and can detect better the targeted grammatical phenomenon when training with a smaller amount of simpler data.
\end{enumerate}

The code and the data are available here: \url{https://github.com/CLCL-Geneva/BLM-SNFDisentangling}.

\paragraph{Terminology}

Sentence embeddings can be read from the output of BERT systems as a $1\times N$  vector (N usually 768 or 1024). This can be viewed as the projection of the sentence into an N-dimensional space. In this paper, we use the word {\it dimensions} to refer to the shape of the data structure used to represent the sentence embeddings. In particular, we use {\it one-dimensional array} to refer to the $1\times N$ vector sentence representation obtained directly from BERT, and {\it 2D representations} to refer to the 2D reshaped ($Rows \times Columns$) array. 

\section{Related work}

Producing sentence representations is a non-trivial issue, mainly because of the structural grammatical and semantic relations they express and their varying complexity and length \cite{stevenson-merlo2022}. The deep learning framework has allowed for a variety of elegant solutions to explicitly learn sentence representations or to induce them as a side-effect or modeling of a more complex problem \citep{mikolov-etal2013,Pennington2014,bojanowski-ea2017enriching,Peters:2018}. Transformer architectures, such as BERT \cite{devlin-etal-2019-bert}, have provided one such solution, where the representation of an input text as a one-dimensional array (usually 1x768 or 1x1024 for the large versions of the model) can be readily obtained from the output of the model. Depending how the system is trained, the sentence embedding can be obtained from the encoding of the [CLS] token, or as a combination of the embeddings of the tokens in the sentence. 

Sentence transformers \cite{reimers-gurevych-2019-sentence} implement architecture and training regime changes on BERT to optimize sentence embeddings for downstream tasks. \newcite{nikolaev-pado-2023-representation} analyze the relation between specific sentence properties (e.g. the contribution of different POS) and the geometry of the embedding space of sentence transformers. 

Whether obtained from sentence transformers or directly from the output of a BERT-based system, sentence embeddings have been shown to  capture information about syntactic and semantic properties. For example, \citet{Manning2020EmergentLS} show that attention heads capture information about dependency relations in transformer models, and \newcite{thrush-etal2020} show the BERT representations contain important information about argument structure and the meaning of verbs. 

Subject-verb agreement is one of the phenomena used to probe a deep-learning system's syntactic abilities. While it is a simple word-level phenomenon, it encodes long-distance relations between words and requires knowledge of structural sentence properties to be correctly learned. 
\newcite{goldberg2019} shows that sentence embeddings capture this property, by testing the language model learned by BERT in predicting the contextually appropriate form of the verb. \citet{linzen2021} include an overview of work that analyzes deep-learning models on this task. While the models tested show high performance in predicting the contextually correct form of a verb, they are guided -- and misled -- by biases within the corpus on which they were trained, e.g. they pay undue attention to the first noun in the sentence. \citeauthor{linzen2021} also include a survey of work that probe deep learning models to understand how grammatical information is encoded. \citet{giulianelli2018,conneau2018,mccoy2018} show that specific grammatical information -- such as the plurality of the subject, the maximal depth of the parse tree of the sentence, the verb auxiliaries -- can be decoded from the sentence encodings (or the hidden state) of the respective systems. \citet{lakretz2021} analyze the actual architecture of an LSTM language model \cite{gulordava2018}, and track the impact of each unit on the long-distance agreement performance. They uncover a combination of a sparse mechanism -- two units -- and a larger distributed circuit that together keep track of number and syntactic structure.

\newcite{lasri-etal-2022-probing} focus on how BERT encodes grammatical number in English and how this information is used for performing number agreement. The focus is on word embeddings and quantifying how much number information they encode at various layers of the BERT architecture. Using a combination of probing approaches, they discover that subjects and predicates embeddings do encode number information, but at different layers. Further investigations into where and how the number information is shared reveals that number information is not directly shared, but rather passed through intermediate tokens.

We also target BERT embeddings to investigate the subject-verb agreement property. Rather than looking at properties of word/token embeddings, we analyze sentence embeddings as the embeddings of the special [CLS] token. We investigate how accessible the number agreement is in raw BERT sentence embeddings in several steps:

\begin{itemize}
    \item test whether the subject-verb agreement rule can be recovered through the sentence representation
    \item test whether different shapes of the sentence embedding -- 1D and various 2D forms -- make the targeted rule more easy to find
    \item test these different shapes of sentence embeddings with several encode-decoder architectures, based on variational autoencoders \cite{kingma2013vae}.
\end{itemize}

\section{Sentence representations for detecting subject-verb agreement}

\subsection{Data}
\label{sec:data}

\begin{figure}[h]
\small
\begin{tabular}{lllll} 
\hline
\multicolumn{5}{c}{\sc Contexts Template}\\
\hline
1 & \textcolor{blue}{NP-sg}& \textcolor{blue}{PP1-sg}& & \textcolor{blue}{VP-sg}  \\
2 & \textcolor{red}{NP-pl} & \textcolor{blue}{PP1-sg}& & \textcolor{red}{VP-pl}  \\
3 & \textcolor{blue}{NP-sg}& \textcolor{red}{PP1-pl} & & \textcolor{blue}{VP-sg}  \\
4 & \textcolor{red}{NP-pl} & \textcolor{red}{PP1-pl} & & \textcolor{red}{VP-pl}  \\
5 & \textcolor{blue}{NP-sg}& \textcolor{blue}{PP1-sg}& \textcolor{blue}{PP2-sg} &\textcolor{blue}{VP-sg}  \\
6 & \textcolor{red}{NP-pl} & \textcolor{blue}{PP1-sg}  & \textcolor{blue}{PP2-sg} &            \textcolor{red}{VP-pl}  \\
7 & \textcolor{blue}{NP-sg}   & \textcolor{red}{PP1-pl} & \textcolor{blue}{PP2-sg} &             \textcolor{blue}{VP-sg}  \\
8 &  \textcolor{red}{NP-pl} & \textcolor{red}{PP1-pl} & \textcolor{blue}{PP2-sg} &               \textcolor{red}{VP-pl}  
\end{tabular}

  \begin{tabular}{rll} \hline
 \multicolumn{3}{c}{{\sc Answer set} } \\ \hline
 1 & NP-sg PP1-sg et NP2 VP-sg & Coord  \\ 
 2 & {\bf NP-pl PP1-pl NP2-sg VP-pl } & correct \\
 3 & NP-sg PP1-sg VP-sg & WNA\\
 4 & NP-sg PP1-sg PP2-sg VP-pl & AE \\
 5 & NP-pl PP1-sg PP1-sg VP-pl & WN1 \\
 6 & NP-pl PP1-pl PP2-pl VP-pl & WN2 \\\hline
 \end{tabular}
\caption{BLM instances for verb-subject agreement, with two attractors. 
WNA=wrong nr. of attractors; AE=agreement error; WN1=wrong nr. for 1$^{st}$ attractor (N1); WN2=wrong nr. for 2$^{nd}$ attractor (N2). }
\label{fig:template-matrices}
\end{figure}

 \begin{figure}[h]
  \small
  \setlength{\tabcolsep}{1mm}
    \begin{tabular}{lllll} 
    \hline
    \multicolumn{5}{c}{\sc Example of  contexts}\\
        \hline
        1 & The vase  & with the flower  &  & leaks. \\
        2 & The vases & with the flower  &  & leak.\\
        3 & The vase  & with the flowers &  & leaks. \\
        4 & The vases & with the flowers &  & leak.\\
        5 & The vase  & with the flower  & from the garden & leaks. \\
        6 & The vases & with the flower  & from the garden & leak.\\
        7 & The vase  & with the flowers & from the garden & leaks. \\
        8 & ???
    \end{tabular}
    
  \setlength{\tabcolsep}{1mm}
    \begin{tabular}{ll} 
    \hline
    \multicolumn{2}{c}{\sc Example of  answers}\\
        \hline
    The vase with the flower  and the garden  leaks. & Coord\\
    \textbf{The vases with the flowers  from the garden  leak.} & Correct\\
    The vase with the flower   leaks. & WNA \\
    The vase with the flower   from the garden  leak. & AE\\
    The vases with the flower  from the garden  leak. & WN1\\
    The vases with the flowers  from the gardens  leak. & WN2\\
    \hline 
    \end{tabular}
 \caption{Examples of actual sentences of type I data (original in French). 
   }
    \label{BLM-agreement}
\end{figure}

Specific grammatical phenomena are often studied on specifically designed or selected datasets (e.g. \cite{nikolaev-pado-2023-representation,linzen2016}). We use BLM-AgrF \cite{an-etal-2023-blm}. The structure of each problem in this task and dataset is inspired from RPM visual pattern tests -- Raven Progressive Matrices -- where one problem consists of overlapping rules the solver must detect \cite{raven1938,chi2019RAVEN}. A Blackbird Language Matrix (BLM) problem \cite{merlo2021} for subject-verb agreement consists of a context set of seven sentences that share the subject-verb agreement phenomenon, but differ in other aspects -- e.g. number of intervening attractors between the subject and the verb, different grammatical numbers for these attractors, and different clause structures. An example template is illustrated in Figure \ref{fig:template-matrices}, and an actual example in Figure \ref{BLM-agreement}. 

The dataset comprises three subsets, of increasing lexical complexity. Type I data is generated based on manually provided seeds, and a template that captures the rules mentioned above. Type II data is generated based on Type I data, by introducing lexical variation with the aid of a transformer, by generating alternatives for masked nouns. Type III data is generated by combining sentences from different instances from the Type II data. This will allow us to investigate the impact of lexical variation on the ability of a system to detect grammatical patterns.

Each subset contains an equal number of instances comprising three clause structures (we include complete instances -- in French -- in Appendix \ref{app:examples}). These structural variations alter the distance and relative depth of the subject and verb to produce a variety of conditions, to allow us to investigate how the subject-verb agreement information is encoded in BERT sentence embeddings.

Each problem is paired with a set of candidate answers. To allow for probing the learned model, apart from the correct answer, the answer sets contain negative examples built by corrupting some of the generating rules. This helps investigate the kind of information and structure learned, and the type of mistakes a system is prone to. 

Table \ref{tab:data} shows the data statistics. Each of the three subsets of datasets is split 90:10 into train and test subsets, which are provided with the data. We use 20\% of the train data for development.

\begin{table}[h]
    \begin{tabular}{lrr} \hline
    dataset & number of problems & train:test split \\ \hline
    Type I  & 2304  & 90:10 \\
    Type II & 38400 & 90:10 \\
    Type III & 38400 & 90:10 \\ \hline
    \end{tabular}
    \caption{Data statistics. The different types of data reflect different amounts of lexical variation within a problem instance.}
    \label{tab:data}
\end{table}

\subsection{Sentence representations}
\label{sec:sentence}

We investigate BERT sentence representations in a series of architectures designed to test whether we can access the relevant information for subject-verb agreement detection. We obtain the sentence embedding from the last layer of BERT, as the embedding of the [CLS] special token.

\begin{figure}[h]
    \begin{tabular}{c} \\ \hline
    {\small FFNN baseline} \\
    \includegraphics[width=0.17\textwidth]{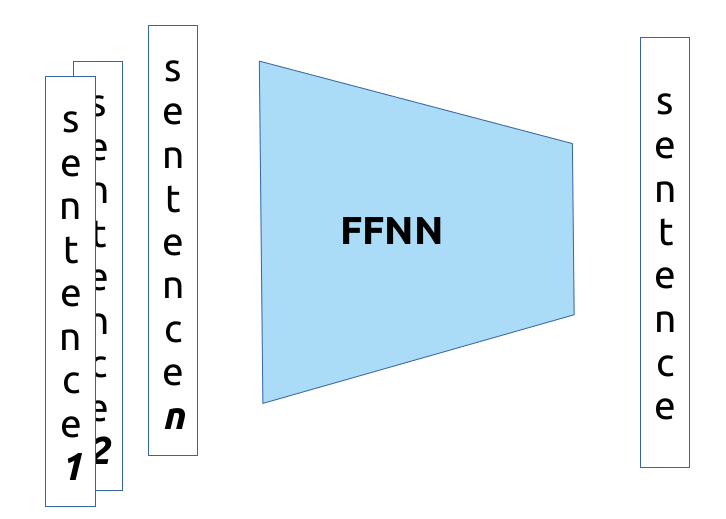} \\ \hline
    {\small CNN baseline} \\
    \includegraphics[width=0.3\textwidth]{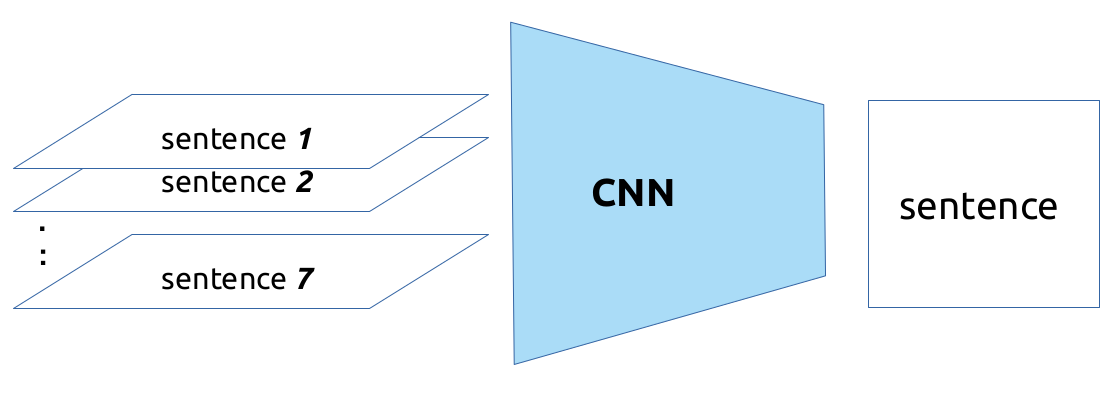} \\ \hline
    {\small encoder-decoder} \\
    \includegraphics[width=0.45\textwidth]{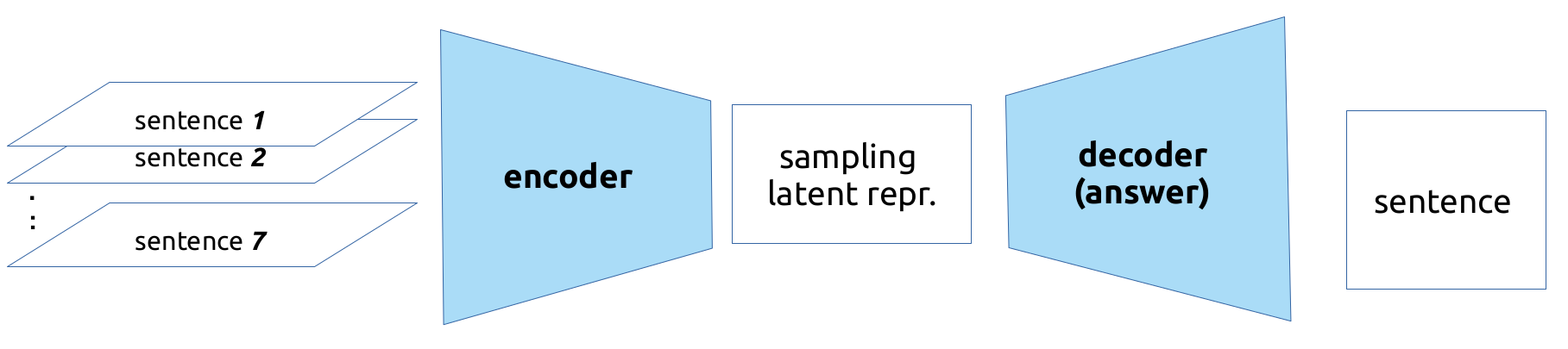} \\ \hline
    {\small dual VAE} \\
    \includegraphics[width=0.45\textwidth]{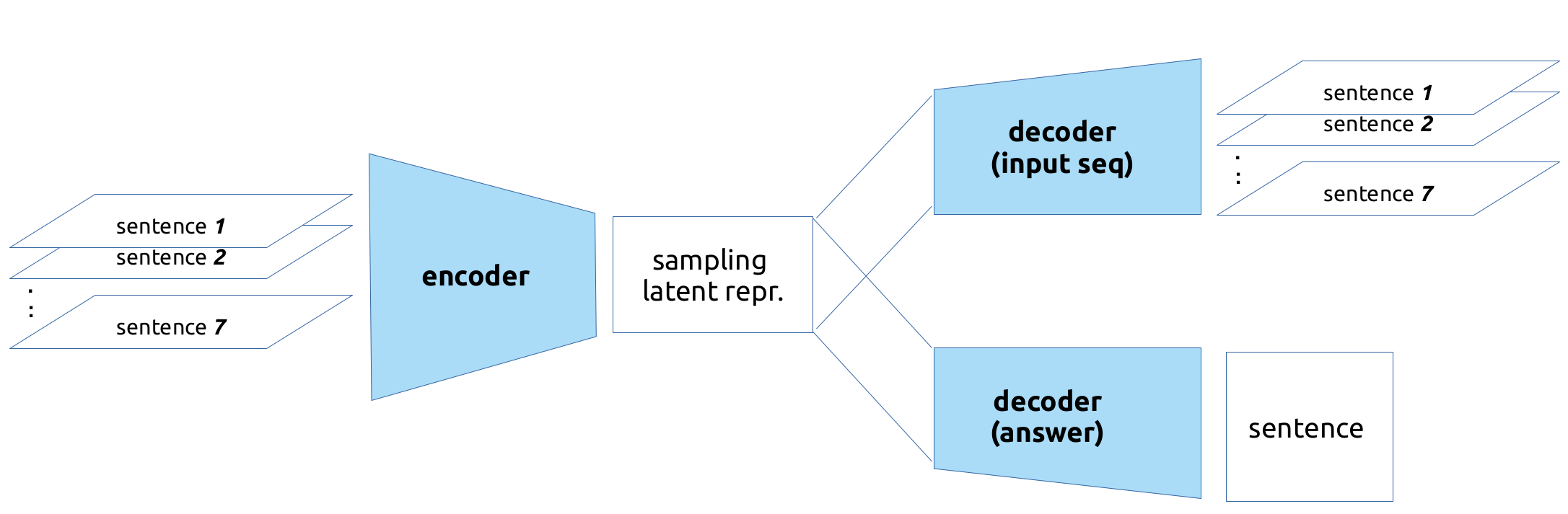} \\ \hline
    \end{tabular}
    \caption{Architecture variations for exploring sentence embeddings}
    \label{fig:architectures}
\end{figure}

Figure \ref{fig:architectures} shows the summary of the architectures explored. Details of the architecture parameters are in Appendix \ref{app:sys_details}. 

To investigate the impact of 2D-ing the sentence embeddings, the input sequences are given as a stack of 1D or 2D-ed sentence embeddings. This sequence of architectures also allows us to test the impact of additional abstracting steps -- through compression into a latent VAE layer -- for accessing patterns that encode the desired information.\footnote{The code is available here: \url{https://github.com/CLCL-Geneva/BLM-SNFDisentangling}}

\paragraph{FFNN baseline}
The FFNN baseline is a three-layer feed-forward neural network. It transforms the sequence of the seven context sentence embeddings into a 1D-tensor, which is passed through three fully-connected layers, and outputs a vector that we take to represent the embedding of the answer sentence. This architecture allows the system to find patterns within and across sentences through the nodes in the successive layers.

The learning objective is to maximize the probability of the correct answer from the candidate answer set. Because the incorrect answers in the answer set are specifically designed to be minimally different from the correct answer, we implement the objective through the max-margin loss function. This function combines the scores of the correct and erroneous ones relative to the sentence embedding predicted by the system. We first compute a score for the embedding $e_i$ of each candidate answer $a_i$ in the answer set $\mathcal{A}$ with respect to the predicted sentence embedding $e_{pred}$ as the cosine of the angle between the respective vectors:

\[
  score(e_i, e_{pred}) = cos(e_i,e_{pred})
\]

The loss uses the max-margin between the score for the correct answer $e_c$ and for each of the incorrect answers $e_i$:

\[
\mathcal{L}_a = \sum_{e_i} [1 - score(e_c, e_{pred}) + score(e_i,e_{pred})]^{+}
\]

For prediction, the answer with the highest $score$ from a candidate set is taken as the correct answer.

\paragraph{CNN baseline}

The CNN baseline consists of N convolutional steps, followed by a linear layer to compress the output to the desired dimensions. The input consists of a stack of sentence representations. We use two variations of this architecture: (i) Baseline\_CNN\_1DxSeq: for a stack of 1D sentence representations, there are three 2D convolutional steps, which use kernels of size 3x3; (ii) Baseline\_CNN\_\{NxM\}: for a stack of (NxM) 2D-ed sentence representations, there is one 3D convolutional layer, with a kernel size 3x15x15. The size of the kernels was set after preliminary experiments. For both variations, the kernels allow the system to detect patterns within a sentence representation and across the sequence.

This system uses the same max-margin loss function as the FFNN baseline system.

\paragraph{Encoder-decoder}

This system is essentially a variational autoencoder (VAE) \cite{kingma2013vae,kingma2015}, but it does not reconstruct the input, rather it constructs an answer. For each of the input variations, the encoder consists of a similar architecture as the corresponding CNN baseline\footnote{Because the FFNN baseline performed very well, and this set-up provides a full receptive field, we also had an encoder-decoder variation using FFNN but it performed much worse than the other variations, and we do not report on it here.}, but the output of the linear layer is the size $L$ of the latent layer ($2 \times L$ to represent the mean and standard deviation for a vector of length $L$). A new vector (of size $L$) is sampled using the output of the encoder as the means and standard deviation of $L$ normal distributions, using the reparametrization trick \cite{kingma2015}. This vector is then unpacked through a decoder to produce a sentence representation. The architecture of the decoder mirrors as close as possible the architecture of the encoder. It differs in that it outputs a single sentence representation, and not the representation of a sequence of  sentences. The two variations are named VAE\_1DxSeq and VAE\_\{MxN\} in the upcoming results tables.

The training objective is to maximize the probability of the correct answer, while improving the approximation of the posterior distribution on the latent layer. This is implemented through a loss function that combines the max-margin loss on the constructed answer (as for previous architectures), with an additional factor -- the regularization factor on the latent layer, typical of VAE models:

\[ \mathcal{L}_l = KL(q_{enc}(z|x)\parallel p(z)) \]

\noindent where $q_{enc}$ is the approximate posterior distribution of $p$, $p(z) = \mathcal{N}(0,1)$ and $q_{enc}(z|x) = \mathcal{N}(\mu_x, \sigma_x)$, where the $[\mu_x; \sigma_x]$ is the latent vector output by the encoder for the input vector $x$. 

The final loss function is:

\[  \mathcal{L}_{enc-dec} = \mathcal{L}_a + \beta * \mathcal{L}_l \]

The $\beta$ coefficient is used to push for disentanglement on the latent layer of a VAE \cite{higgins2016beta}. For the reported experiments $\beta = 1$.

\paragraph{Dual VAE}

The dual VAE adds a decoder to reconstruct the input to the encoder-decoder system, which  mirrors the encoder in architecture and parameters. The two variations are Dual\_VAE\_1DxSeq and Dual\_VAE\_\{NxM\}.

The training objective is to maximize both the probability of the answer and the reconstructed input, and improve the approximation of the posterior distribution on the latent layer. Essentially, we add a factor to the loss function of the encoder-decoder architecture, reflecting the reconstruction loss:

\[
    \mathcal{L}_{dVAE} = \mathcal{L}_a + \alpha * \mathcal{L}_{recon} + \beta * \mathcal{L}_l
\]

\noindent where $\alpha$ is a coefficient that can control the contribution of the input reconstruction signal. For the experiments reported, $\alpha = 0.01$, and serves as a scaling factor, to bring the value of the reconstruction loss within the same magnitude as the answer loss $\mathcal{L}_a$. The reconstruction loss is computed as the mean-square error between the representation of the input sequence ($x$) and the reconstructed one ($x'$): $\mathcal{L}_{recon} = MSE(x,x') $

\begin{figure*}[t]
    \includegraphics[width=\textwidth]{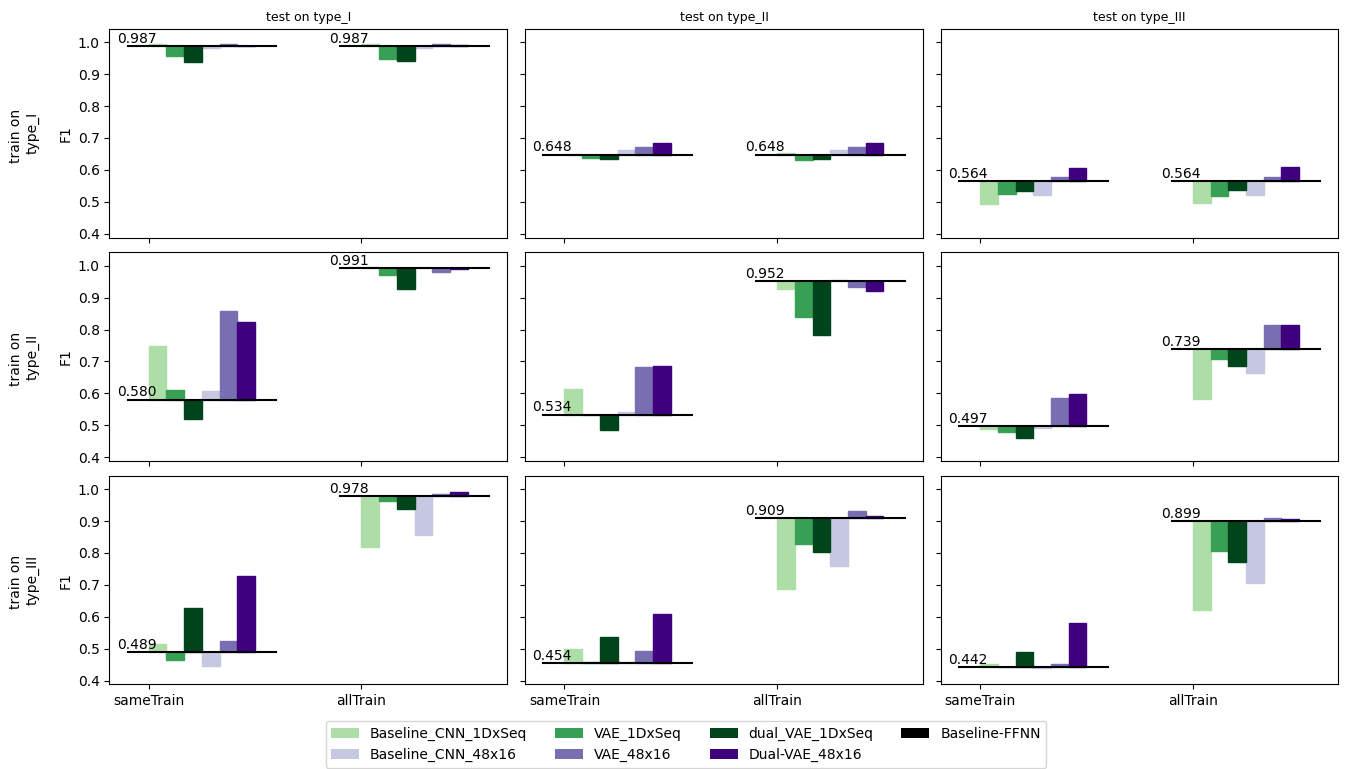}
    \caption{(best seen in color) F1 scores (averages over five runs) on the FFNN baseline and CNN, enc-dec and dual VAE systems with 48x16 2D-ed sentence embeddings. The graphs shows the difference in F1 between the systems relative to the reference baseline (FFNN). Each panel includes results on training on the same amount of data (left) and training on full data (right).}
    \label{fig:results_sys_comparison}
\end{figure*}

\section{Experiments}

We report experiments on seven systems -- the baseline FFNN, and two variations (the input as a stack of 1D or 2D sentence representations) for each of the other three architectures.

Our aim was to explore raw BERT sentence embeddings, and not variations fine-tuned for specific tasks. The results reported here are based on sentence embeddings obtained using BERTTokenizer and BERTModel from the transformers Python library, using the pretrained "BERT-base-multilingual-cased" model\footnote{\url{https://huggingface.co/bert-base-multilingual-cased}}.\footnote{ We have run preliminary experiments with French-specific sentence embeddings using FlauBERT \cite{le-etal-2020-flaubert}. The results were lower than when using a multilingual cased BERT language model.} 

Preliminary experiments have been used to determine the kernel size for processing the 2D and 3D tensors with the stack of 1D and 2D-ed sentence embeddings respectively. The optimal kernel for 2D tensors was 3x3, and for the 3D tensors was 3x15x15. This large kernel size imposes specific restrictions on the dimensions of the 2D-ed sentence embeddings. Because a sentence embedding is a one-dimensional array of length 768, there are only 4 possible 2D values, where both dimensions are greater than 15 (16x48, 24x32, 32x24, 48x16). 

We have also explored the size of the latent layer (5:25, step 5), and chosen the latent size 5. 
All systems used a learning rate of 0.001 and Adam optimizer, and batch size 100. For the baselines and experiments on the full training set for type II and type III data the training was done for 50 epochs, and for type I data and the sameTrain set-up, the training was done for 120 epochs.

The experiments were run on an HP PAIR Workstation Z4 G4 MT, with an Intel Xeon W-2255 processor, 64G RAM, and a MSI GeForce RTX 3090 VENTUS 3X OC 24G GDDR6X GPU.

\begin{figure*}[t]
    \includegraphics[width=\textwidth]{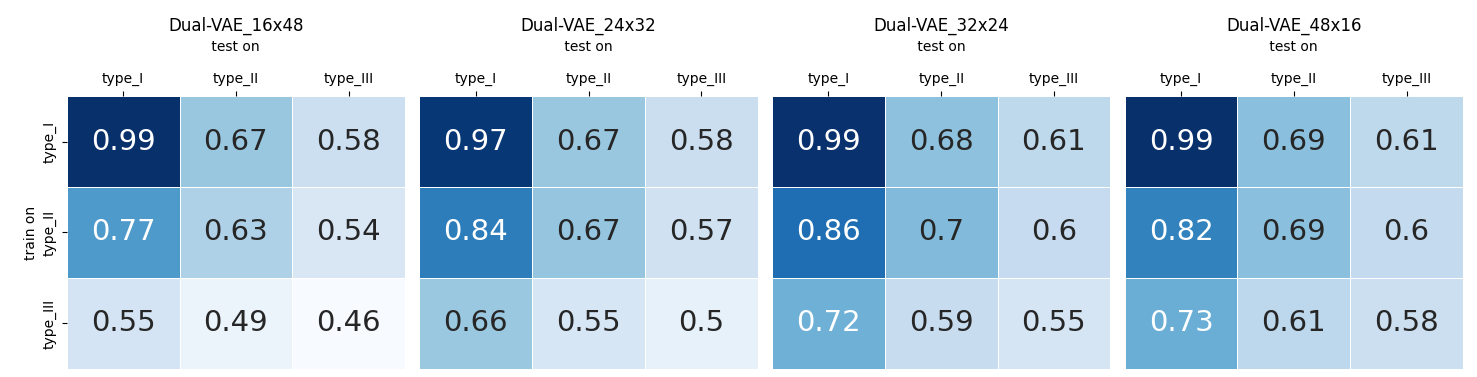}
    \caption{Impact of different reshapings: F1 results (averages over five runs) on (16x48), (24x32), (32x24), (48x16) reshaping, using the dual VAE architecture, and trained on similar amounts of training data.}
    \label{fig:results_2ded}
\end{figure*}

\subsection{Results}

Figures \ref{fig:results_sys_comparison}  and \ref{fig:results_2ded} show the main experimental results. All results represent F1 averages over five runs. More details are provided in Appendix \ref{app:results}. 

Figure \ref{fig:results_sys_comparison} shows the difference in F1 between the FFNN baseline (the black lines) and the increasingly complex architectures. Each row corresponds to the data type used for training, and the columns to the data type used for testing. Each panel shows the test results of models trained on the same amount of data, on the left, and training on full data, on the right.

Figure \ref{fig:results_2ded} shows the results obtained using different 2D transformations of the one-dimensional tensor BERT sentence embeddings with the dual VAE system, when training on the same amount of data (2073 instances -- the amount available for \ti data).   
Overall, the best-performing embedding is the one reshaped in a 48x16 matrix, the setting then used for the results reported in the system study, shown in Figure \ref{fig:results_sys_comparison}.

Figure \ref{fig:training_data} shows the impact of the amount of training data on the performance of the models. The results reported are average F1 scores over 5 runs, using the dual-VAE architecture with 48x16 sentence embedding.

\subsection{Discussion}

\paragraph{Impact of 2D-ed representations and VAE-based architectures}

In Figure \ref{fig:results_sys_comparison}, the horizontal black lines represent the performance of the Baseline FFNN system, and the bars show the relative performance of the system variations with 1D and 2D-ed sentence representations.


When using the full training data the results on \tii and \tiii subsets are very high. This is in line with ML theory, as input with more variety leads to better-performing models, {\it when given enough training data}. The low results of the baselines and the systems using the 1D representations on the restricted training set-up (2073 instances -- the available amount of training+validation data for \ti -- for all subsets) shows that these systems do not access the most relevant information from the sentence embedding for our targeted phenomenon.

Pairwise comparisons of similar architectures with different types of input show that 2D-ed representations lead to better results in almost all settings, particularly in the harsher training scenario with limited data (the left bar group in each plot). 

The progression of architectures -- from the CNN to the dual VAE -- also show an increase in results, for both types of input representations. The phenomenon is more evident -- and more useful -- particularly for the restricted training scenario. It shows that forcing the representations to more compressed and abstract forms is useful for distilling the information useful to detect our targeted grammatical phenomenon.

The most interesting result is a combination of the impact of the 2D-ed representations and the various architectures: as the panels corresponding to training on \ti data (first row in Figure \ref{fig:results_sys_comparison}) show, the combination of the dual VAE architecture with 2D-ed sentence embedding leads to the best results when testing on \tii and \tiii data, which are lexically more complex than \ti data. This shows that with a good combination of input representation and system, a model can find robust patterns even in a smaller amount of simple data.

\begin{figure*}[t]
    \includegraphics[width=\textwidth]{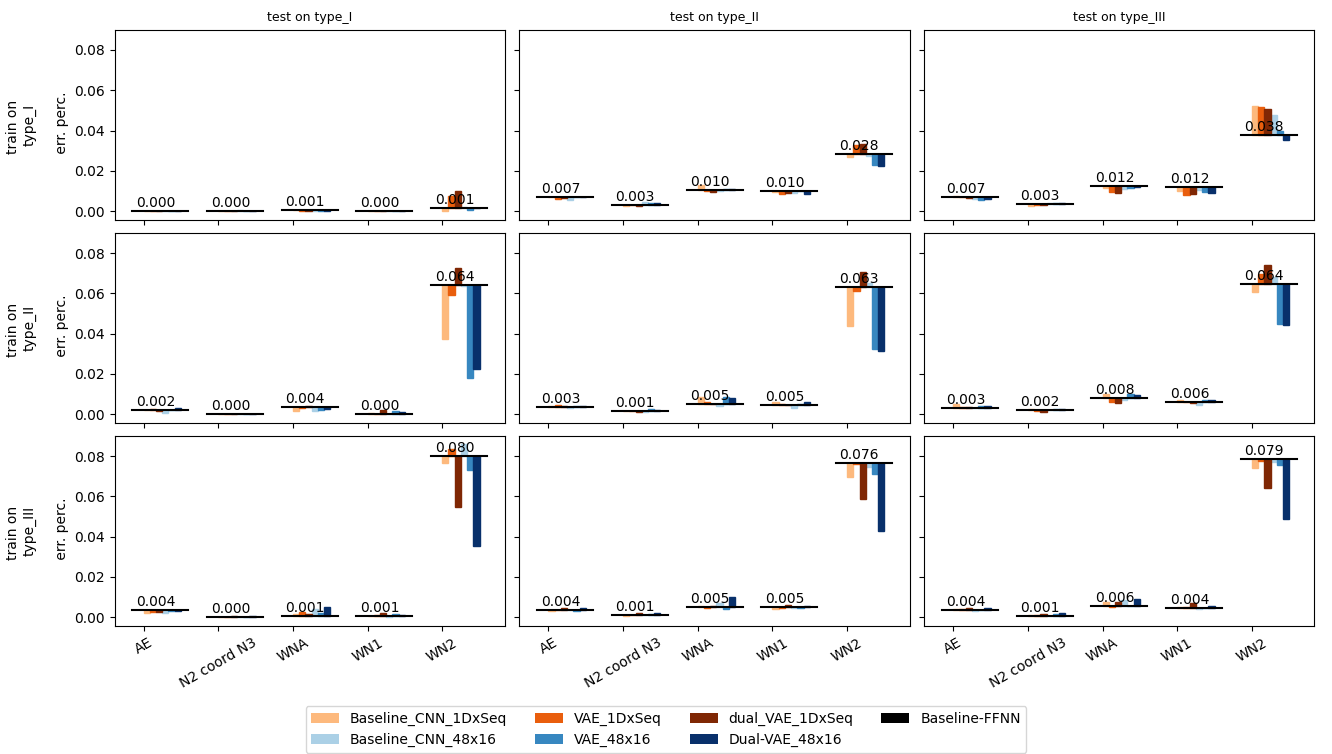}
    \caption{(best seen in color) Error analysis on the four systems trained with the same amount of data. The y-axis is the percentage of the error relative to the size of the test set (i.e. downward bars indicate an improvement.)}
    \label{fig:error_analysis}
\end{figure*}

\paragraph{Impact of 2D-ing the sentence representation} The results presented in Figure \ref{fig:results_sys_comparison} show that the 48x16 2D version of the sentence representation leads to better results than the 1x768 version. Figure \ref{fig:results_2ded} shows the impact of various 2D variations, when training the systems on the same amount of training+validation data. The results are obtained with the dual VAE architecture. The (overall) best performing representation from the 4 variations is the 48x16 version. In fact, for the $N \times M$ 2D-ing, the results are better the smaller $M$ becomes. This indicates that information is somehow uniformly distributed in a BERT sentence embedding, within subsequences of length close to 16 -- at least the information relevant to our particular subject-verb agreement task.

\begin{figure}[h!]
    \centering
    \includegraphics[width=0.45\textwidth]{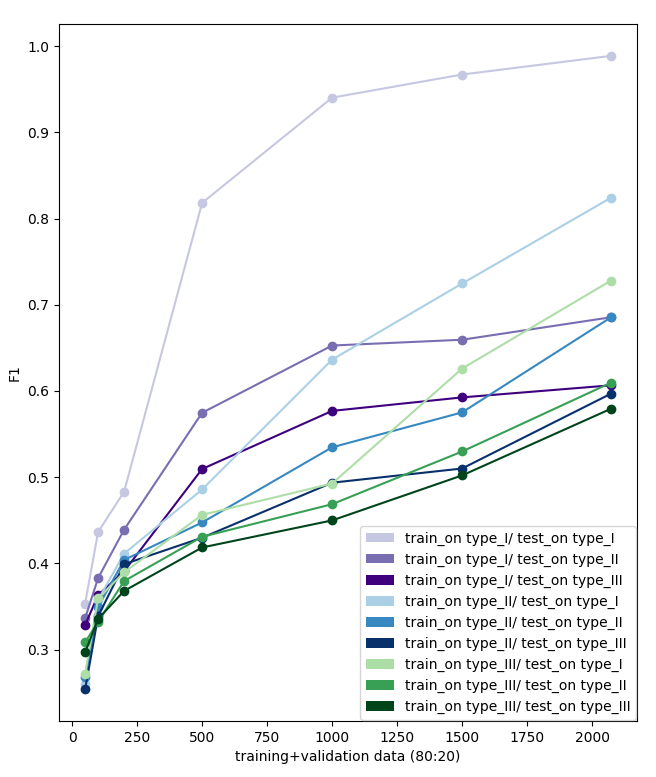}
    \vspace{-3mm}
    \caption{(best seen in color) Training data analysis using the the Dual\_VAE\_48x16 system}
    \label{fig:training_data}
\end{figure}

\paragraph{Impact of training data} Figure \ref{fig:results_sys_comparison} shows that the 2D-ed sentence representation combined with the dual VAE architecture leads to the best results, particularly when training the systems on the same amount of training+validation data. We further analyze the learning curves when varying the amount of training+validation data from 50 to 2073 (split 80:20 into training and validation data). Figure \ref{fig:training_data} shows these results. 

The curves corresponding to training on \ti data -- in shades of purple -- approach the higher performance fastest, showing that they are able to exploit smaller amounts of data better. The lexical variations in \tii and \tiii data seem to obfuscate the targeted patterns, as they require more training data.

\subsection{Error Analyses}

The error analysis shown in Figure \ref{fig:error_analysis} clearly shows that errors are almost always of the same kind ---the assignment of the wrong grammatical number to the second attractor, closer to the verb--- thus performing a local kind of agreement instead of the correct longer-distance structural agreement. These types of errors are frequent in humans too \cite{linzen-leonard2018}.
It can be seen that the dual-VAE corrects these errors to a great degree (and greater than the other models), thus showing that it can detect the non-local patterns better than the other architectures. The combination of 2D-ing the sentence embedding and the large size of the kernel allows the system to find more distant patterns in the sentence embedding, thus connecting more distant tokens.

\section{Conclusion}

We have proposed and investigated variations of BERT sentence representations for the task of identifying the rule of subject-verb agreement and some of its properties in a dataset consisting of sequences of sentences that instantiate these underlying rules. BERT sentence embeddings, as 1D arrays, have been successfully used for a variety of tasks, but the information they encode about specific phenomena is distributed over the vector. Reshaping the sentence representation into two-dimensional inputs leads to improved results, and an additional step of abstracting these 2D-ed sentence embeddings leads to further improvements for the task. These more abstracted 2D-ed embeddings also learn a robust model based on a smaller amount of simpler training data, while showing good performance on the more lexically-complex data.

We plan to explore whether BERT sentence embedding may have even higher-dimension patterns, and whether such nD-ed BERT sentence representation can be used to detect other grammatical or semantic phenomena. We also plan to directly distill 2D sentence representation in more compact and disentangled representation, to encode more explicitly some of the distributed information in these embeddings.

\section*{Limitations}

The experiments reported were performed on a dataset of French sentences, with a particular organization: sequences of sentences as input, each with a slightly different structure but sharing the subject-verb agreement rule. All sentences in the input sequence are processed together. In future work we plan to separate the distillation of rules from a sentence representation from the processing of the sequence. 

We have investigated only part of the parameters in the proposed architectures. In particular, the $\beta$ coefficient in the encoder-decoder and the dual VAE architectures was set to 1. Higher values may lead to more disentangled representations on the latent layer.

\section*{Ethics Statement}
To the best of our knowledge, there are no ethics concerns with this paper. 

\section*{Acknowledgments}

We gratefully acknowledge the partial support of this work by the Swiss National Science Foundation, through grants  \#51NF40\_180888 (NCCR Evolving Language) and SNF Advanced grant  TMAG-1\_209426 to PM. 
\medskip

\bibliography{anthology,custom}
\bibliographystyle{acl_natbib}

\newpage
\appendix

\onecolumn
\section{Supplementary Materials}
\vspace{-1.2cm}

\subsection{Type I instance examples}
\label{app:examples}

\begin{figure}[h]
\small
{
\begin{tabularx}{\textwidth}{rllllll} 
\hline
\multicolumn{7}{l}{Main clause}                                                                                                                                                               \\ 
\hline
1 &                  & \textcolor{blue}{L'ordinateur}   & \textcolor{blue}{avec le programme}  &                                    &                      & \textcolor{blue}{est en panne.}  \\
2 &                  & \textcolor{red}{Les ordinateurs} & \textcolor{blue}{avec le programme}  &                                    &                      & \textcolor{red}{sont en panne.}  \\
3 &                  & \textcolor{blue}{L'ordinateur}   & \textcolor{red}{avec les programmes} &                                    &                      & \textcolor{blue}{est en panne.}  \\
4 &                  & \textcolor{red}{Les ordinateurs} & \textcolor{red}{avec les programmes} &                                    &                      & \textcolor{red}{sont en panne.}  \\
5 &                  & \textcolor{blue}{L'ordinateur}   & \textcolor{blue}{avec le programme}  & \textcolor{blue}{de l'exp\'{e}rience} &                      & \textcolor{blue}{est en panne.}  \\
6 &                  & \textcolor{red}{Les ordinateurs} & \textcolor{blue}{avec le programme}  & \textcolor{blue}{de l'exp\'{e}rience} &                      & \textcolor{red}{sont en panne.}  \\
7 &                  & \textcolor{blue}{L'ordinateur}   & \textcolor{red}{avec les programmes} & \textcolor{blue}{de l'exp\'{e}rience} &                      & \textcolor{blue}{est en panne.}  \\
8 &                  & \textcolor{red}{Les ordinateurs} & \textcolor{red}{avec les programmes} & \textcolor{blue}{de l'exp\'{e}rience} &                      & \textcolor{red}{sont en panne.}  \\ 
\hline\hline
\multicolumn{7}{l}{Completive clause}                                                                                                                                                         \\ 
\hline
1 & Jean suppose que & \textcolor{blue}{l'ordinateur}   & \textcolor{blue}{avec le programme}  &                                    &                      & \textcolor{blue}{est en panne.}  \\
2 & Jean suppose que & \textcolor{red}{les ordinateurs} & \textcolor{blue}{avec le programme}  &                                    &                      & \textcolor{red}{sont en panne.}  \\
3 & Jean suppose que & \textcolor{blue}{l'ordinateur}   & \textcolor{red}{avec les programmes} &                                    &                      & \textcolor{blue}{est en panne.}  \\
4 & Jean suppose que & \textcolor{red}{les ordinateurs} & \textcolor{red}{avec les programmes} &                                    &                      & \textcolor{red}{sont en panne.}  \\
5 & Jean suppose que & \textcolor{blue}{l'ordinateur}   & \textcolor{blue}{avec le programme}  & \textcolor{blue}{de l'exp\'{e}rience} &                      & \textcolor{blue}{est en panne.}  \\
6 & Jean suppose que & \textcolor{red}{les ordinateurs} & \textcolor{blue}{avec le programme}  & \textcolor{blue}{de l'exp\'{e}rience} &                      & \textcolor{red}{sont en panne.}  \\
7 & Jean suppose que & \textcolor{blue}{l'ordinateur}   & \textcolor{red}{avec les programmes} & \textcolor{blue}{de l'exp\'{e}rience} &                      & \textcolor{blue}{est en panne.}  \\
8 & Jean suppose que & \textcolor{red}{les ordinateurs} & \textcolor{red}{avec les programmes} & \textcolor{blue}{de l'exp\'{e}rience} &                      & \textcolor{red}{sont en panne.}  \\ 
\hline\hline
\multicolumn{7}{l}{Relative clause}                                                                                                                                                           \\ 
\hline
1 &                  & \textcolor{blue}{L'ordinateur}   & \textcolor{blue}{avec le programme}  &                                    & dont Jean se servait & \textcolor{blue}{est en panne.}  \\
2 &                  & \textcolor{red}{Les ordinateurs} & \textcolor{blue}{avec le programme}  &                                    & dont Jean se servait & \textcolor{red}{sont en panne.}  \\
3 &                  & \textcolor{blue}{L'ordinateur}   & \textcolor{red}{avec les programmes} &                                    & dont Jean se servait & \textcolor{blue}{est en panne.}  \\
4 &                  & \textcolor{red}{Les ordinateurs} & \textcolor{red}{avec les programmes} &                                    & dont Jean se servait & \textcolor{red}{sont en panne.}  \\
5 &                  & \textcolor{blue}{L'ordinateur}   & \textcolor{blue}{avec le programme}  & \textcolor{blue}{de l'exp\'{e}rience} & dont Jean se servait & \textcolor{blue}{est en panne.}  \\
6 &                  & \textcolor{red}{Les ordinateurs} & \textcolor{blue}{avec le programme}  & \textcolor{blue}{de l'exp\'{e}rience} & dont Jean se servait & \textcolor{red}{sont en panne.}  \\
7 &                  & \textcolor{blue}{L'ordinateur}   & \textcolor{red}{avec les programmes} & \textcolor{blue}{de l'exp\'{e}rience} & dont Jean se servait & \textcolor{blue}{est en panne.}  \\
8 &                  & \textcolor{red}{Les ordinateurs} & \textcolor{red}{avec les programmes} & \textcolor{blue}{de l'exp\'{e}rience} & dont Jean se servait & \textcolor{red}{sont en panne.}  \\
\hline
\end{tabularx}
 \\ \\ \\
  \begin{tabular}{rll} \hline
 \multicolumn{3}{c}{{\bf Answer set} for problem constructed from lines 1-7 of the main clause sequence} \\ \hline
 1 & L'ordinateur avec le programme et l'experi\'{e}nce est en panne. & N2 coord N3 \\ 
 2 & {\bf Les ordinateurs avec les programmes de l'experi\'{e}nce sont en panne.} & correct \\
 3 & L'ordinateur avec le programme est en panne. & wrong number of attractors \\
 4 & L'ordinateur avec les programmes de l'experi\'{e}nce sont en panne. & agreement error \\
 5 & Les ordinateurs avec le programme de l'experi\'{e}nce sont en panne. & wrong nr. for 1$^{st}$ attractor noun (N1) \\
 6 & Les ordinateurs avec les programmes des experi\'{e}nces sont en panne. & wrong nr. for 2$^{nd}$ attractor noun (N2) \\ \hline
 \end{tabular}

}
\caption{BLM-AgrF instances for verb-subject agreement, with two attractors (programme, experi\'{e}nce), and three clause structures. And candidate answer set for a problem constructed from lines 1-7 of the main clause sequence.}
\label{fig:matrices}
\end{figure}

\subsection{System architecture details}
\label{app:sys_details}

\begin{minipage}[t]{\textwidth}
{\bf Baseline\_FFNN } 
\begin{verbatim}
===================================================================================
Layer (type:depth-idx)                   Output Shape              Param #
===================================================================================
BaselineFFNN                             [100, 768]                --
--Linear: 1-1                            [100, 1536]               8,259,072
--Linear: 1-2                            [100, 1536]               2,360,832
--Linear: 1-3                            [100, 768]                1,180,416
===================================================================================
Total params: 11,800,320
Trainable params: 11,800,320
Non-trainable params: 0
Total mult-adds (G): 1.18
===================================================================================
Input size (MB): 2.15
Forward/backward pass size (MB): 3.07
Params size (MB): 47.20
Estimated Total Size (MB): 52.42
===================================================================================   
\end{verbatim} 
\end{minipage}

\begin{minipage}[t]{\textwidth}
{\bf Baseline\_CNN\_1DxSeq}: CNN with stack of 1D sentence embeddings 
\begin{verbatim}
===================================================================================
Layer (type:depth-idx)                   Output Shape              Param #
===================================================================================
BaselineCNN_1DxSeq                       [100, 768]                --
--Conv2d: 1-1                            [100, 4, 5, 766]          40
--Conv2d: 1-2                            [100, 8, 3, 764]          296
--Conv2d: 1-3                            [100, 16, 1, 762]         1,168
--Linear: 1-4                            [100, 768]                9,364,224
===================================================================================
Total params: 9,365,728
Trainable params: 9,365,728
Non-trainable params: 0
Total mult-adds (G): 1.11
===================================================================================
Input size (MB): 2.15
Forward/backward pass size (MB): 37.29
Params size (MB): 37.46
Estimated Total Size (MB): 76.91
===================================================================================
\end{verbatim} 
\end{minipage}

\begin{minipage}[t]{\textwidth}
{\bf Baseline\_CNN\_48x16}: CNN with 48x16 2D-ed sentence embedding
\begin{verbatim}
===================================================================================
Layer (type:depth-idx)                   Output Shape              Param #
===================================================================================
BaselineCNN                              [100, 7, 768]                --
--Conv3d: 1-1                            [100, 32, 5, 34, 2]       21,632
--Linear: 1-2                            [100, 768]                8,356,608
===================================================================================
Total params: 8,378,240
Trainable params: 8,378,240
Non-trainable params: 0
Total mult-adds (G): 1.57
===================================================================================
Input size (MB): 2.15
Forward/backward pass size (MB): 9.32
Params size (MB): 33.51
Estimated Total Size (MB): 44.98
===================================================================================
\end{verbatim} 
\end{minipage}

\begin{minipage}[t]{\textwidth}
{\bf VAE\_1DxSeq}: encoder-decoder with stack of 1D sentence embeddings 
\begin{verbatim}
===================================================================================
Layer (type:depth-idx)                   Output Shape              Param #
===================================================================================
VariationalAutoencoder                   [100, 5]                  --
--Encoder: 1-1                           [100, 5]                  --
     --Conv2d: 2-1                       [100, 4, 5, 766]          40
     --Conv2d: 2-2                       [100, 8, 3, 764]          296
     --Conv2d: 2-3                       [100, 16, 1, 762]         1,168
     --Linear: 2-4                       [100, 10]                 121,930
--simpleSampling: 1-2                    [100, 5]                  --
--Decoder_answer: 1-3                    [100, 1, 1, 768]          --
     --Linear: 2-5                       [100, 12192]              73,152
     --ConvTranspose2d: 2-6              [100, 8, 1, 764]          392
     --ConvTranspose2d: 2-7              [100, 4, 1, 766]          100
     --ConvTranspose2d: 2-8              [100, 1, 1, 768]          13
===================================================================================
Total params: 197,091
Trainable params: 197,091
Non-trainable params: 0
Total mult-adds (M): 230.28
===================================================================================
Input size (MB): 2.15
Forward/backward pass size (MB): 54.40
Params size (MB): 0.79
Estimated Total Size (MB): 57.33
===================================================================================
\end{verbatim} 
\end{minipage}

\begin{minipage}[t]{\textwidth}
{\bf VAE\_48x16}: encoder-decoder with 48x16 2D-ed sentence embeddings \\
\begin{verbatim}
===================================================================================
Layer (type:depth-idx)                   Output Shape              Param #
===================================================================================
VariationalAutoencoder                   [100, 5]                  --
--Encoder: 1-1                           [100, 5]                  --
     --Conv3d: 2-1                       [100, 32, 5, 34, 2]       21,632
     --Linear: 2-2                       [100, 10]                 108,810
--simpleSampling: 1-2                    [100, 5]                  --
--Decoder_answer: 1-3                    [100, 1, 1, 48, 16]       --
     --Linear: 2-3                       [100, 2176]               13,056
     --ConvTranspose3d: 2-4              [100, 1, 1, 48, 16]       7,201
===================================================================================
Total params: 150,699
Trainable params: 150,699
Non-trainable params: 0
Total mult-adds (G): 1.30
===================================================================================
Input size (MB): 2.15
Forward/backward pass size (MB): 11.07
Params size (MB): 0.60
Estimated Total Size (MB): 13.82
===================================================================================
\end{verbatim} 
\end{minipage}

\begin{minipage}[t]{\textwidth}
{\bf Dual\_VAE\_1DxSeq}: dual VAE with stack of 1D sentence embeddings \\
\begin{verbatim}
===================================================================================
Layer (type:depth-idx)                   Output Shape              Param #
===================================================================================
VariationalAutoencoder                   [100, 5]                  --
--Encoder: 1-1                           [100, 5]                  --
     --Conv2d: 2-1                       [100, 4, 5, 766]          40
     --Conv2d: 2-2                       [100, 8, 3, 764]          296
     --Conv2d: 2-3                       [100, 16, 1, 762]         1,168
     --Linear: 2-4                       [100, 10]                 121,930
--simpleSampling: 1-2                    [100, 5]                  --
--Decoder_mirror: 1-3                    [100, 1, 7, 768]          --
     --Linear: 2-5                       [100, 12192]              73,152
     --ConvTranspose2d: 2-6              [100, 8, 3, 764]          1,160
     --ConvTranspose2d: 2-7              [100, 4, 5, 766]          292
     --ConvTranspose2d: 2-8              [100, 1, 7, 768]          37
--Decoder_answer: 1-4                    [100, 1, 1, 768]          --
     --Linear: 2-9                       [100, 12192]              73,152
     --ConvTranspose2d: 2-10             [100, 8, 1, 764]          392
     --ConvTranspose2d: 2-11             [100, 4, 1, 766]          100
     --ConvTranspose2d: 2-12             [100, 1, 1, 768]          13
===================================================================================
Total params: 271,732
Trainable params: 271,732
Non-trainable params: 0
Total mult-adds (M): 635.19
===================================================================================
Input size (MB): 2.15
Forward/backward pass size (MB): 95.37
Params size (MB): 1.09
Estimated Total Size (MB): 98.61
===================================================================================
\end{verbatim} 
\end{minipage}

\begin{minipage}[t]{\textwidth}
{\bf Dual\_Vae\_48x16}: dual VAE with 48x16 2D-ed sentence embeddings \\
\begin{verbatim}
===================================================================================
Layer (type:depth-idx)                   Output Shape              Param #
===================================================================================
VariationalAutoencoder                   [100, 5]                  --
--Encoder: 1-1                           [100, 5]                  --
     --Conv3d: 2-1                       [100, 32, 5, 34, 2]       21,632
     --Linear: 2-2                       [100, 10]                 108,810
--simpleSampling: 1-2                    [100, 5]                  --
--Decoder_mirror: 1-3                    [100, 1, 7, 48, 16]       --
     --Linear: 2-3                       [100, 10880]              65,280
     --ConvTranspose3d: 2-4              [100, 1, 7, 48, 16]       21,601
--Decoder_answer: 1-4                    [100, 1, 1, 48, 16]       --
     --Linear: 2-5                       [100, 2176]               13,056
     --ConvTranspose3d: 2-6              [100, 1, 1, 48, 16]       7,201
===================================================================================
Total params: 237,580
Trainable params: 237,580
Non-trainable params: 0
Total mult-adds (G): 12.92
===================================================================================
Input size (MB): 2.15
Forward/backward pass size (MB): 24.07
Params size (MB): 0.95
Estimated Total Size (MB): 27.17
===================================================================================
\end{verbatim} 
\end{minipage}

\newpage

\subsection{Detailed experimental results}
\label{app:results}

\paragraph{Results on 2D-ing BERt sentence embeddings}

The detailed version of the results in Figure \ref{fig:results_2ded}
\vspace{-5mm}
\begin{table}[h]
    \centering
    \begin{tabular}{llrrrr} \\ 
    {\sc train on} & {\sc test on} & {\sc 16x48 F1 (std)} & {\sc 24x32 F1 (std)} & {\sc 32x24 F1 (std)} & {\sc 48x16 F1 (std)} \\ \hline
\ti   &   \ti & {\bf 0.9905} (0.0069) & 0.9740 (0.0061) & 0.9879 (0.0064) & 0.9887 (0.0080) \\ \cline{3-6}
\ti   &  \tii & 0.6682 (0.0035) & 0.6683 (0.0041) & 0.6815 (0.0021) & {\bf 0.6855} (0.0028) \\ \cline{3-6}
\ti   & \tiii & 0.5795 (0.0049) & 0.5819 (0.0011) & {\bf 0.6072} (0.0045) & 0.6066 (0.0018) \\ \cline{2-6}
\tii  &   \ti & 0.7732 (0.0283) & 0.8355  (0.0137) & {\bf 0.8632} (0.0089) & 0.8242 (0.0065) \\ \cline{3-6}
\tii  &  \tii & 0.6333 (0.0132) & 0.6725 (0.0077) & {\bf 0.6984} (0.0046) & 0.6855 (0.0040) \\ \cline{3-6}
\tii  & \tiii & 0.5431 (0.0069) & 0.5689 (0.0024) & 0.5952 (0.0058) & {\bf 0.5969} (0.0046) \\ \cline{2-6}
\tiii &   \ti & 0.5550 (0.0461) & 0.6649 (0.0059) & 0.7221 (0.0228) & {\bf 0.7281} (0.0259) \\ \cline{3-6}
\tiii &  \tii & 0.4947 (0.0218) & 0.5474 (0.0076) & 0.5884 (0.0053) & {\bf 0.6096} (0.0064) \\ \cline{3-6}
\tiii & \tiii & 0.4620 (0.0151) & 0.5042 (0.0089) & 0.5515 (0.0050) & {\bf 0.5794} (0.0054) \\ \hline   
    \end{tabular}
\caption{Analysis of 2D-ing the BERT sentence embeddings: F1 (std) scores for the four 2D combinations. The highest value for each train/test combination highlighted in bold.}
\label{tab:2ding}
\end{table}

\paragraph{Results on system analysis}

The detailed version of the results in  Figure \ref{fig:results_sys_comparison}

\begin{table}[h]
    \vspace{-5mm}
    \centering
    \begin{tabular}{llrp{2.5cm}p{2.5cm}p{2.5cm}} \\ 
{\sc train on} & {\sc test on} & {\sc Baseline\_FFNN} & {\sc Baseline\_CNN 48x16} & {\sc VAE\_48x16} & {\sc Dual\_VAE\_48x16} \\ \hline
\multicolumn{6}{l}{{\sc train on full training data}} \\ \hline
\ti   &   \ti & 0.9870 (0) & 0.9827 (0) & {\bf 0.9957} (0) & 0.9905 (0.0042) \\ \cline{3-6}
\ti   &  \tii & 0.6482 (0) & 0.6612 (0) & 0.6729 (0) & {\bf 0.6829} (0.0070) \\ \cline{3-6}
\ti   & \tiii & 0.5643 (0) & 0.5229 (0) & 0.5776 (0) & {\bf 0.6089} (0.0062) \\ \cline{2-6}
\tii  &   \ti & {\bf 0.9913} (0) & {\bf 0.9913} (0) & 0.9801 (0.0052) & 0.9896 (0.0044) \\ \cline{3-6}
\tii  &  \tii & 0.9523 (0) & {\bf 0.9552} (0) & 0.9331 (0.0014) & 0.9215 (0.0017) \\ \cline{3-6}
\tii  & \tiii & 0.7391 (0) & 0.6622 (0) & {\bf 0.8156} (0.0019) & 0.8152 (0.0037) \\ \cline{2-6}
\tiii &   \ti & 0.9784 (0) & 0.8571 (0) & 0.9853 (0.0035) & {\bf 0.9913} (0.0027) \\ \cline{3-6}
\tiii &  \tii & 0.9086 (0) & 0.7578 (0) & {\bf 0.9309} (0.0039) & 0.9146 (0.0040) \\ \cline{3-6}
\tiii & \tiii & 0.8987 (0) & 0.7062 (0) & {\bf 0.9089} (0.0016) & 0.9047 (0.0046) \\ \hline
\multicolumn{6}{l}{{\sc train on the same amount of data (2073 instances: 1658 train/415 validation)}} \\ \hline
\ti   &   \ti & 0.9870 (0) & 0.9827 (0) & {\bf 0.9957} (0) & 0.9887 (0.0080) \\ \cline{3-6}
\ti   &  \tii & 0.6482 (0) & 0.6612 (0) & 0.6729 (0) & {\bf 0.6855} (0.0028) \\ \cline{3-6}
\ti   & \tiii & 0.5643 (0) & 0.5229 (0) & 0.5776 (0) & {\bf 0.6066} (0.0018) \\ \cline{2-6}
\tii  &   \ti & 0.5801 (0) & 0.6061 (0) & {\bf 0.8571} (0) & 0.8242 (0.0065) \\ \cline{3-6}
\tii  &  \tii & 0.5336 (0) & 0.5411 (0) & 0.6833 (0) & {\bf 0.6855} (0.0040) \\ \cline{3-6}
\tii  & \tiii & 0.4974 (0) & 0.4901 (0) & 0.5846 (0) & {\bf 0.5969} (0.0046) \\ \cline{2-6}
\tiii &   \ti & 0.4892 (0) & 0.4459 (0) & 0.5238 (0) & {\bf 0.7281} (0.0259) \\ \cline{3-6}
\tiii &  \tii & 0.4542 (0) & 0.4576 (0) & 0.4935 (0) & {\bf 0.6096} (0.0064) \\ \cline{3-6}
\tiii & \tiii & 0.4419 (0) & 0.4380 (0) & 0.4529 (0) & {\bf 0.5794} (0.0054) \\ \hline
    \end{tabular}
\caption{Analysis of systems: F1 (std) scores for the FFNN baseline and the 3 2D-ed system architectures. The highest value for each train/test combination is highlighted in bold.}
\label{tab:sys_analysis_2d}
\end{table}

\begin{table}
    \centering
    \begin{tabular}{llccc} \\ 
{\sc train on} & {\sc test on} & {\sc Baseline\_CNN\_1DxSeq} & {\sc VAE\_1DxSeq} & {\sc Dual\_VAE\_1DxSeq} \\ \hline
\multicolumn{5}{l}{{\sc train on full training data}} \\ \hline
\ti   &   \ti &  0.9948	(0.0064) & 0.9489 (0.0088) & 0.9403 (0.0084) \\ \cline{3-5} 
\ti   &  \tii &  0.6518	(0.0044) & 0.6305 (0.0069) & 0.6347	(0.0026) \\ \cline{3-5}
\ti   & \tiii &  0.4961	(0.0043) & 0.5194 (0.0086) & 0.5357	(0.0037) \\ \cline{2-5}
\tii  &   \ti &  0.9974	(0.0035) & 0.9697 (0.0039) & 0.9264 (0.0152) \\ \cline{3-5}
\tii  &  \tii &  0.9256	(0.0041) & 0.8393 (0.0016) & 0.7819 (0.0111) \\ \cline{3-5}
\tii  & \tiii &  0.5837	(0.0064) & 0.7060 (0.0154) & 0.6853	(0.0107) \\ \cline{2-5}
\tiii &   \ti &  0.8173	(0.0390) & 0.9628 (0.0065) & 0.9385 (0.0050) \\ \cline{3-5}
\tiii &  \tii &  0.6865	(0.0191) & 0.8279 (0.0060) & 0.8020	(0.0020) \\ \cline{3-5}
\tiii & \tiii &  0.6205	(0.0180) & 0.8046 (0.0038) & 0.7727 (0.0036) \\ \hline
\multicolumn{5}{l}{{\sc train on the same amount of data (2073 instances: 1658 train/415 validation)}} \\ \hline
\ti   &   \ti &  0.9939	(0.0044) & 0.9558 (0.0042) & 0.9385	(0.0069) \\ \cline{3-5}
\ti   &  \tii &  0.6491	(0.0070) & 0.6358 (0.0039) & 0.6335	(0.0018) \\ \cline{3-5}
\ti   & \tiii &  0.4946	(0.0047) & 0.5249 (0.0092) & 0.5346	(0.0051) \\ \cline{2-5}
\tii  &   \ti &  0.7489	(0.0387) & 0.6095 (0.0202) & 0.5203 (0.0330) \\ \cline{3-5}
\tii  &  \tii &  0.6146	(0.0180) & 0.5306 (0.0077) & 0.4862	(0.0185) \\ \cline{3-5}
\tii  & \tiii &  0.4898	(0.0062) & 0.4774 (0.0048) & 0.4608 (0.0072) \\ \cline{2-5}
\tiii &   \ti &  0.5134	(0.0193) & 0.4641 (0.0213) & 0.6286	(0.1111) \\ \cline{3-5}
\tiii &  \tii &  0.4994	(0.0093) & 0.4595 (0.0071) & 0.5377	(0.0502) \\ \cline{3-5}
\tiii & \tiii &  0.4516	(0.0019) & 0.4440 (0.0051) & 0.4884	(0.0299) \\ \hline
    \end{tabular}
\caption{Analysis of systems: F1 (std) scores for the three 1DxSeq system architectures. The highest value for each train/test combination highlighted in bold.}
\label{tab:sys_analysis_1d}
\end{table}

\end{document}